\documentclass{MIPRO}

\usepackage{cite}
\usepackage{amsmath,amssymb,amsfonts}
\usepackage{algorithmic}
\usepackage{graphicx}
\usepackage{textcomp}
\usepackage{xcolor}
\usepackage[T1]{fontenc}
\usepackage{flushend}
\usepackage{multirow}
\usepackage{multicol}
\usepackage{array}
\usepackage{url}
\usepackage{booktabs}
\usepackage{tabularx}

\begin{document}

\title{Large Language Models as Optimizers: A Survey of Direct vs. Tool-Augmented Approaches and Their Performance Frontiers}

\author{
\IEEEauthorblockN{
Roko Peran\IEEEauthorrefmark{1},
Luka Hobor\IEEEauthorrefmark{2},
Mihael Kovac\IEEEauthorrefmark{2},
Mario Brcic\IEEEauthorrefmark{2} 
}

\IEEEauthorblockA{\IEEEauthorrefmark{1}  
Student at University of Zagreb Faculty of Electrical Engineering and Computing, Zagreb, Croatia }

\IEEEauthorblockA{\IEEEauthorrefmark{2}  
University of Zagreb Faculty of Electrical Engineering and Computing, Zagreb, Croatia }
roko.peran@fer.hr, luka.hobor@fer.hr, mihael.kovac@fer.hr, mario.brcic@fer.hr
}
\maketitle

\begin{abstract}
Large Language Models (LLMs) are increasingly involved in complex mathematical optimization, even if the pragmatic user who triggers them is unaware of it. After all, many real-world problems reduce to the search for better or the best solutions. The field of LLM-as-optimizer has three paradigms: direct optimization, tool-augmented optimization, and tool-creating optimization. Direct optimization uses iterative prompting and heuristic generation to navigate solution spaces. Tool-augmented optimization translates natural language problems into formal specifications and orchestrates external solvers. Tool-creating optimization goes further, using LLMs to discover reusable algorithms or heuristics that can be deployed at zero marginal LLM cost. We describe current performance frontiers based on the benchmarks from the literature. We identify the critical reasoning gap in current architectures and argue for trade-offs between the future potential of direct optimization and the auditability of tool-augmented optimization. Even future, more powerful models might opt for tool-making to improve operational efficiency for repetitive families of problems.
\end{abstract}

\renewcommand\IEEEkeywordsname{Keywords}

\section{Introduction}

Large Language Models (LLMs) are now applied to complex optimization tasks such as vehicle routing, resource scheduling, hyperparameter tuning, and algorithm discovery \cite{yang2024opro, romera2024funsearch, ahmaditeshnizi2024optimus, daros2025llmco_survey}.

Many real-world optimization problems start as natural-language descriptions — delivery constraints in prose, experimental goals in a lab notebook. Traditional optimization requires translating these into formal mathematical programs before a solver (e.g., Gurobi, CPLEX, Z3) can be used. LLMs can bridge this gap by solving problems directly or by automating the translation.

Three paradigms have emerged. In \textit{direct optimization}, the LLM itself acts as the optimizer: generating candidate solutions, receiving feedback, and iteratively refining them \cite{yang2024opro, iklassov2024sge}. In \textit{tool-augmented optimization}, the LLM serves as a translator and orchestrator, converting natural language problems into formal mathematical programs and dispatching them to external solvers \cite{ahmaditeshnizi2024optimus, jiang2025llmopt}. In \textit{tool-creating} optimization, LLMs generate reusable heuristic algorithms that run within solver or metaheuristic frameworks rather than formulating individual problems \cite{romera2024funsearch, ye2024reevo}.

This survey reviews these three paradigms, reports performance frontiers from the literature, and identifies the reasoning gap that limits direct approaches. Section~II covers direct optimization. Section~III covers tool-augmented methods. Section~IV describes tool-creating approaches. Section~V analyzes why LLMs work as optimizers, the role of model choice, and where they fail. Section~VI discusses open challenges.


\section{Direct Optimization}

The direct paradigm asks: \textit{can an LLM navigate a solution space without any external solver?} The LLM generates candidate solutions, receives feedback on quality, and iteratively refines them. The paradigm has grown from small prompting tasks to NP-hard combinatorial optimization, but a fundamental scale limit remains.

\subsection{Foundational Arc: From Prompting to Iterative Search (2022--2024)}

Chain-of-Thought (CoT, 2022) \cite{wei2022cot} and Tree-of-Thought (ToT, 2023) \cite{yao2023tot} showed that intermediate reasoning steps improve LLM performance on structured tasks. \textbf{OPRO} (2024) \cite{yang2024opro} built on this by framing optimization as a natural language loop: the LLM receives past solutions with their objective values and generates improved candidates. Using GPT-4 and PaLM 2, OPRO achieved prompt optimization gains of +8\% on GSM8K and +50\% on BBH, but performance degrades beyond approximately 20 variables, where simple heuristics outperform all tested LLMs on TSP. \textbf{SGE} (2024) \cite{iklassov2024sge} extended iterative prompting to NP-hard combinatorial problems by decomposing them into subtasks and exploring multiple reasoning paths (+27.8\% over prior prompting methods). Related approaches applied the same principle to planning (\textbf{AdaPlanner}, 2023 \cite{sun2023adaplanner}) and reinforcement learning without gradient updates \cite{brooks2023incontext}.

\subsection{Expanding to New Problem Classes (2023--2024)}

The direct paradigm generalizes well where LLM domain knowledge provides an advantage over classical optimizers. In \textit{hyperparameter optimization}, Zhang et al.\ (2023) \cite{zhang2023llmhpo} showed GPT-4-guided search competitive with Bayesian optimization, exploiting the model's ability to propose architectural changes beyond numerical search spaces. \textbf{LLAMBO} (2024) \cite{liu2024llambo} formalized this by replacing the Gaussian Process surrogate with an LLM, achieving strong early-search efficiency---e.g., chemical synthesis yield from 25.2\% to 60.7\%. For \textit{scheduling}, Abgaryan et al.\ \cite{abgaryan2024schedule} fine-tuned an LLM on 120k job-shop instances, reaching performance comparable to graph neural networks with reinforcement learning. Liu et al.\ \cite{liu2023llm4moea} applied LLMs as crossover and mutation operators in multi-objective evolutionary optimization.

\subsection{The Fine-Tuning Frontier (2025)}

The current frontier replaces prompting-time reasoning with training-time internalization. \textbf{E2E CO Solver} (2025) \cite{jiang2025e2eco} trains a 7B-parameter open-weight LLM via supervised fine-tuning and feasibility-and-optimality-aware reinforcement learning, mapping natural language descriptions directly to solutions across seven NP-hard problem types with 1--8\% optimality gaps, without any solver or code generation. LLaMoCo \cite{ma2024llamoco} similarly instruction-tunes open-weight models for optimization code generation. These approaches represent a qualitative shift: the model internalizes optimization heuristics rather than reasoning from scratch.

The direct paradigm is strongest for \textit{ill-structured problems}, where formal constraint specification is impractical, and for \textit{expensive-evaluation settings}, where domain knowledge accelerates search. Fine-tuning pushes scale limits, but the reasoning gap at higher complexity has not been overcome by prompting alone (Section~V).

\section{Tool-Augmented Optimization}\label{sec:toolaug}

\begin{figure}
  \centering
  \includegraphics[width=0.95\columnwidth]{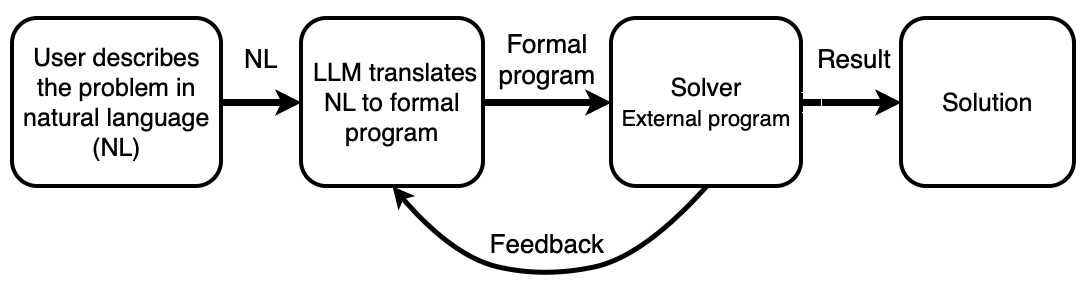}
  \caption{Workflow for tool-augmented approach}
  \label{fig:tool_augmented_approach}
\end{figure}

The tool-augmented paradigm delegates solving to external engines (Figure~\ref{fig:tool_augmented_approach}). This division of labor achieves the highest accuracy on structured benchmarks and inherits the solver's mathematical guarantees when the formulations are correct. Formulation correctness is the central challenge, and the field's evolution since 2023 is largely a story of identifying and attacking successive bottlenecks in that process.

\subsection{From Pipeline to Trained System (2023--2025)}

\textbf{OptiMUS} (2024) \cite{ahmaditeshnizi2024optimus} established the canonical architecture: natural language input $\rightarrow$ LLM formulation $\rightarrow$ solver code $\rightarrow$ execution $\rightarrow$ error feedback. Built on GPT-4, its multi-agent Formulator/Programmer/Evaluator structure outperformed standard prompting by over 40 percentage points on NLP4LP. Two responses attacked different limitations. \textbf{CoE} (2024) \cite{xiao2024coe} added specialized expert agents with backward reflection targeting complex industrial OR. \textbf{ORLM} (2025) \cite{Huang_2025} took a data-centric path, fine-tuning open-weight 7B-scale models (LLaMA-based \cite{touvron2023llama}) on OR-Instruct, a semi-automated data synthesis framework, surpassing GPT-4 prompting by 42.2\%. \textbf{LLMOPT} (2025) \cite{jiang2025llmopt} synthesized both directions with a universal five-element formulation, SFT+DPO alignment, and auto-testing self-correction, reaching 97.3\% on NL4Opt. \textbf{SIRL} (2025) \cite{chen2025sirl} used solver execution outputs as verifiable rewards during reinforcement learning, with SIRL-32B (Qwen-based \cite{yang2025qwen3}) surpassing DeepSeek-V3 and OpenAI-o3 on most benchmarks. Deng et al.\ \cite{deng2024cafa} showed that coding-as-formulation can boost LLM performance on LP without fine-tuning.

Parallel branches expanded the scope beyond LP/MILP. \textbf{SATLM} (2023) \cite{ye2023satlm} applied LLMs to SAT/SMT specifications. Michailidis et al.\ \cite{michailidis2024cpmodelling} demonstrated constraint modeling via in-context learning. Hao et al.\ \cite{hao2025llmfp} used LLM-based formalized programming for zero-shot planning. \textbf{MCP-Solver} (2025) \cite{mcpsolver2025} standardized the solver interface across MiniZinc, PySAT, and Z3. Commercial deployments confirm pipeline viability: NVIDIA cuOpt \cite{nvidia2025cuopt} enabled a 120$\times$ increase in routing speed; OptiGuide \cite{li2023optiguide} handles supply chain queries at Microsoft with 93\% accuracy.

\subsection{The Correctness Frontier (2025--2026)}

Architectural and training advances pushed accuracy near the ceiling on simple benchmarks, revealing a deeper bottleneck: \textit{silent formulation failures}. The standard feedback loop---re-feeding solver error messages to the LLM---addresses only the 21--31\% of failures that are pure coding errors; formulation errors produce code that executes cleanly yet encodes the wrong problem \cite{xiao2025ijcaisurvey}. Self-correction has structural limits: models oscillate between correct and incorrect formulations, or revise already-correct code \cite{zhang2024darkside}.

Three 2025--2026 approaches attack this from different angles. \textbf{SAC-Opt} \cite{zhang2025sacopt} reconstructs semantic anchors to identify and repair logically incorrect formulations (+21.9\% on complex LP). \textbf{Autoformulation} (2025) \cite{astorga2025autoformulation} reframes formulation as a combinatorial search, using MCTS with SMT-based symbolic pruning to explore alternative mathematical models. \textbf{ReLoop} (2026) \cite{reloop} introduces \textit{behavioral verification}: perturbing constraint parameters to test whether the objective responds---if it does not, a constraint is missing. That raises correctness from 22.6\% to 31.1\% for Claude Opus, while revealing the full severity of the problem: DeepSeek-V3.2 achieves 91.1\% solver feasibility but only 0.5\% formulation correctness in compositional scenarios, a 90-point gap.

\subsection{Benchmarks and Performance Frontiers}

The evaluation landscape mirrors the field's open challenge. Table~\ref{tab:benchmarks} summarizes key benchmarks and best-reported results.

\begin{table}[htbp]
\caption{Tool-augmented optimization benchmarks. \textit{Reliability} reflects ground-truth annotation quality: H\,=\,hand-verified; M\,=\,est.\ $\leq$20\% errors; L\,=\,$>$20\% errors (up to 54\% for IndustryOR) \cite{xiao2025ijcaisurvey}.}
\label{tab:benchmarks}
\centering
\resizebox{\columnwidth}{!}{%
\begin{tabular}{lccccc}
\hline
\textbf{Benchmark} & \textbf{Size} & \textbf{Types} & \textbf{Best} & \textbf{System} & \textbf{Rel.} \\
\hline
NL4Opt & 289 & LP & 97.3\% & LLMOPT \cite{jiang2025llmopt} & M \\
NLP4LP & 52--355 & LP, MILP & 89\% & SIRL \cite{chen2025sirl} & M \\
MAMO-Easy & 652 & LP & 96\% & LLMOPT \cite{jiang2025llmopt} & M \\
MAMO-Complex & 211 & LP & 85.8\% & LLMOPT \cite{jiang2025llmopt} & L \\
IndustryOR & 100 & Real-world & 57\% & LLMOPT \cite{jiang2025llmopt} & L \\
RetailOpt-190 & 190 & Compositional & 31.1\% & ReLoop \cite{reloop} & H \\
\hline
\end{tabular}%
}
\end{table}

Near-perfect accuracy on simple LP drops to 57\% on real-world industrial OR and 31.1\% on compositional multi-constraint scenarios. These figures must be read as \textbf{optimistic upper bounds}: \textbf{benchmark annotations contain} \textbf{15--54\% errors} \cite{xiao2025ijcaisurvey}, making observed accuracy partly an artifact of measurement. RetailOpt-190\cite{reloop}, with hand-verified solutions and jointly binding constraint sets, is the first step toward reliable evaluation at realistic complexity.

\section{Tool-Creating Approaches: LLM-Driven Algorithm Discovery}

Tool-creating approaches use LLMs to \textit{discover algorithms}. These algorithms are general and execute within code executors, solvers, or metaheuristic frameworks, rather than addressing just a single problem instance. The critical distinction from tool-augmented optimization lies in breaking the limits imposed by predefined tools. Tool-augmented systems are limited in expressivity and capability by the available toolset. In contrast, tool-creating systems incur a higher upfront LLM cost to generate a \textit{reusable} tool that can be deployed independently, achieving effectively zero marginal LLM cost relative to tool-augmented approaches as the initial investment is amortized over repeated use. The paradigm emerged from 2023--2024 work on evolutionary program search, reached production deployment by 2025, and is the fastest-growing subfield, with a recent survey \cite{liu2024algdesign_survey} cataloging over 180 papers.

\subsection{Evolutionary Program Search}

The foundational work is \textbf{FunSearch} (2024) \cite{romera2024funsearch}, which uses a PaLM 2-based Codey model as a mutation operator within an evolutionary framework to evolve program-space heuristics. FunSearch discovered bin-packing strategies that outperform classical heuristics and constructed large cap sets. Reproducibility is limited: only 4 of 140 independent runs discovered the claimed cap set result, and the approach exhibits path-dependence from the LLM's sensitivity to its own prior outputs.

\textbf{EoH} (2024) \cite{liu2024eoh} addressed a key FunSearch limitation by evolving both natural language \textit{thoughts} and code simultaneously. This dual representation enables more efficient search---EoH discovers superior bin-packing heuristics while requiring significantly fewer LLM calls with GPT-4.

\textbf{AlphaEvolve} (2025) \cite{alphaevolve2025} scales the paradigm to production with a two-tier architecture: Gemini Flash generates high-volume candidates while Gemini Pro handles quality-critical mutations. It discovered a scheduling heuristic deployed on Google's Borg cluster and achieved a 23\% speedup on a Gemini training kernel. These results emerge from massive compute and proprietary infrastructure, making independent replication difficult.

\subsection{Reflective and Agent-Based Approaches}

\textbf{ReEvo} (2024) \cite{ye2024reevo} introduced ``verbal gradients'': the LLM compares two heuristics and articulates improvements in natural language, making evolutionary search more directed. Across six combinatorial optimization problem types, ReEvo generated state-of-the-art heuristics in approximately five minutes, substantially more sample-efficient than FunSearch.

\textbf{VRPAgent} (2025) \cite{hottung2025vrpagent} applied the paradigm to a specialized niche: generating problem-specific destroy/repair operators for Large Neighborhood Search, becoming the first LLM-based approach to advance the state of the art on established vehicle routing benchmarks.

\subsection{Assessment}
The field must contend with several open questions. First, current successes concentrate on problems with fast, reliable fitness evaluation (bin-packing, routing, scheduling); generalization to domains with expensive or noisy evaluation remains untested. Second, the evolutionary search is constrained by the LLM's training distribution; it can recombine and refine patterns it has seen, but the extent to which it can discover genuinely novel algorithmic ideas is debated \cite{liu2024algdesign_survey}.

\section{Successes and Failures of LLMs in Optimization}

\subsection{Mechanisms of Effectiveness}

Two properties underlie LLM effectiveness in optimization. First, \textit{knowledge-driven reasoning}: training on mathematical texts, engineering documents, and code gives LLMs both domain knowledge and a library of solution strategies. That allows them to interpret problem context (e.g., inferring implicit engineering constraints) and recall relevant heuristics (e.g., nearest neighbor for TSP) in ways that pure numerical solvers cannot \cite{jiang2025dsm, huang2024llmopt_survey}. Second, \textit{in-context learning}: when provided with candidate solutions and their objective values, LLMs infer improvement patterns and extrapolate better solutions, learning task-specific heuristics from examples in the prompt \cite{yang2024opro}.

\subsection{The Role of Model Choice}

Not all LLMs are interchangeable in optimization tasks. Three distinctions matter in practice:

\textit{Scale and access.} Prompting-based systems (OPRO, OptiMUS, CoE) typically rely on frontier closed-weight models---GPT-4 \cite{openai2023gpt4}, Claude, or Gemini---whose large parameter counts and extensive training data give them the broadest domain knowledge. Fine-tuning approaches (ORLM, SIRL, E2E CO Solver, LLaMoCo) require open-weight models, typically in the 7B--32B range based on LLaMA \cite{touvron2023llama} or Qwen \cite{yang2025qwen3} families. These smaller models can match or exceed frontier models on specific tasks after domain-specific training, as ORLM-7B surpasses GPT-4 by 42.2

\textit{Reasoning-specialized models.} Models with extended chain-of-thought capabilities---OpenAI o1/o3 \cite{openai2024o1} and DeepSeek-R1 \cite{guo2025deepseek}---reduce formulation errors by $\sim$7\% on complex problems \cite{zhang2025orllmagent}. However, the benefit is not uniform: on simpler benchmarks, reasoning overhead can degrade performance, and for fine-tuned models, CoT can conflict with learned generation templates (Section~V-C).

\textit{Multi-model architectures.} AlphaEvolve demonstrates that combining a lightweight model (Gemini Flash) for volume with a capable model (Gemini Pro) for quality-critical steps outperforms either model alone. The LATM framework \cite{cai2024latm} formalizes a similar principle: a powerful model creates tools that a lightweight model applies.

\subsection{The Reasoning Gap}

Despite these capabilities, LLMs exhibit fundamental limitations. The most systematic evidence comes from three analyses.

\textit{The constraint formulation bottleneck}: automated constraint formulation constitutes the primary performance bottleneck in tool-augmented systems \cite{chen2026optengine}. When TSP instances scale from $N=5$ to $N=8$, frontier models abandon formal formulation entirely, reverting to greedy heuristics.

\textit{Solution-based metrics mask formulation errors}: component-level evaluation reveals that an LLM can achieve a perfect optimality gap while capturing only 50\% of ground-truth constraints, because the relaxed problem happens to produce the same optimal value \cite{componenteval2025}. 

\textit{Systematic error patterns}: recurring errors cluster predictably by problem class. Subtour elimination constraints in TSP are frequently misapplied, flow conservation sign errors recur in network problems, and binary variable declarations are often omitted \cite{optimind2025}.

The most common formulation errors, ranked by prevalence, are: missing or incorrect constraints (especially implicit domain knowledge), wrong variable types, parameter extraction errors, and wrong objective functions. Performance degrades nonlinearly with problem complexity. Shojaee et al.\ \cite{shojaee2025illusion} found ``complete accuracy collapse'' beyond certain complexity thresholds, with reasoning effort peaking then declining despite adequate token budgets---evidence that current LLMs rely on pattern matching rather than genuine algorithmic reasoning.

ReLoop's evaluation shows the feasibility--correctness gap across frontier models: DeepSeek-V3.2 achieves 91.1\% solver-feasibility but only 0.5\% formulation correctness; Claude Opus achieves 72.1\% and 22.6\% respectively \cite{reloop}.

The effect of extended reasoning on this gap is mixed. Reasoning models \cite{guo2025deepseek, openai2024o1} reduce errors by $\sim$7\% on complex problems \cite{zhang2025orllmagent}. Still, CoT can \textit{increase} errors for fine-tuned models: OptMATH-32B's accuracy drops from 56.2\% to 30\% when CoT is applied, because the structured generation template conflicts with the format learned during fine-tuning \cite{reloop}. This tension between prompting-time and training-time strategies is unresolved.

\subsection{Trade-offs Across the Three Paradigms}

Table~\ref{tab:paradigm} summarizes the trade-offs across the three paradigms.

\begin{table}[htbp]
\caption{Comparison of optimization paradigms.}
\label{tab:paradigm}
\centering
\begin{tabularx}{\linewidth}{lXXX}
\toprule
\textbf{Criterion} & \textbf{Direct} & \textbf{Tool-aug.} & \textbf{Tool-creating} \\
\midrule
Quality & Optimal at small scale & Provably optimal & Novel SOTA \\
\midrule
Scale & $\leq$20 vars & Millions of vars & Heuristics scale freely \\
\midrule
LLM cost & High (many calls) & Low (1--10 calls) & High upfront; zero at deploy \\
\midrule
Auditability & Low & High (formal model) & Medium (code artifacts) \\
\midrule
Formal guarantees & None & Solver guarantees (if model correct) & None; heuristic \\
\midrule
Explainability & Low (black-box iterates) & High (model + code auditable) & Medium (algorithm readable) \\
\midrule
Robustness & Low; prompt-sensitive & Medium; solver validates & Medium; tested over instances \\
\midrule
Best for & Prompt opt., small problems & Structured LP/MILP at scale & Algorithm discovery \\
\bottomrule
\end{tabularx}
\end{table}

Tool-augmented optimization provides the strongest \textit{formal guarantees}: when the formulation is correct, the solver certifies optimality. Direct and tool-creating approaches offer no such certificates. \textit{Explainability} follows the same gradient---tool-augmented systems expose auditable mathematical models, tool-creating systems produce inspectable code, and direct systems offer little transparency. \textit{Robustness} is highest in tool-augmented settings, where the solver rejects invalid formulations, though feasible-but-incorrect models remain a blind spot (Section~\ref{sec:toolaug}). Direct approaches are sensitive to prompt phrasing \cite{mirzadeh2025gsmsymbolic}.

\section{Open Challenges and Future Directions}

\subsection{Benchmark Quality}

The field cannot reliably measure progress when benchmark error rates reach 15--54\% \cite{xiao2025ijcaisurvey}. 

\subsection{The Complexity Cliff}

Current LLMs exhibit sharp performance degradation beyond certain problem-complexity thresholds. This ``complexity cliff'' suggests reliance on pattern matching to training examples rather than genuine mathematical reasoning. Closing the gap between 97\% accuracy on 5-variable LPs and reliable performance on industrial-scale problems remains the main open challenge.

\subsection{Emerging Techniques}

Three directions address the complexity cliff. \textit{Extended-reasoning models} \cite{guo2025deepseek, openai2024o1} outperform standard LLMs by at least 7\% on complex problems, though with mixed results on simpler benchmarks \cite{zhang2025orllmagent}. \textit{Search-based formulation} treats the mathematical model as a search target: SolverLLM \cite{chen2025solvellm} and Autoformulation \cite{astorga2025autoformulation} use MCTS to explore formulation alternatives, outperforming single-generation approaches without fine-tuning. \textit{Solver-grounded RL training} (SIRL \cite{chen2025sirl}, E2E CO Solver \cite{jiang2025e2eco}) uses solver execution outcomes as training signals, so the model learns optimization patterns that prompting alone cannot teach.

\subsection{Reproducibility Crisis in Tool-Creating Approaches}

The tool-creation paradigm faces a reproducibility crisis that the community has not yet systematically addressed. FunSearch's headline cap-set result was discovered in only 4 of 140 independent runs; AlphaEvolve's production results are entirely inaccessible to independent researchers due to proprietary infrastructure. Yet both are cited as SOTA results and influence the direction of follow-on work.

How should the community handle benchmark claims from systems that cannot be independently replicated? We propose three norms: benchmark claims from closed systems should be labeled as such, open systems should report result distributions rather than best-case outcomes, and fitness evaluators should be treated as reproducible artifacts.

\subsection{Tool-Making as a Meta-Strategy}

The LATM framework \cite{cai2024latm} establishes the principle of LLMs creating reusable tools: a powerful model creates utility functions that a lightweight model then applies, dramatically reducing inference cost. This pattern manifests in optimization through reusable formulation templates. Even future, more powerful models might opt for tool-making over direct solving, since creating a reusable solver component amortizes the LLM's reasoning cost across an entire family of problems.


\section{Conclusion}

This survey has examined three paradigms for LLM-based optimization. Direct optimization is broadening from prompt optimization to end-to-end combinatorial solving but remains limited by reasoning capacity at scale. Tool-augmented optimization achieves strong results on structured benchmarks but is bottlenecked by available tools and formulation accuracy on complex, compositional problems. Tool-creating approaches have produced notable results in algorithm discovery, though reproducibility and generalization remain open questions. The choice between paradigms depends on model capabilities: frontier closed-weight models excel at prompting-based formulation, fine-tuned open-weight models can surpass them on specific tasks, and reasoning-specialized models reduce but do not eliminate the complexity cliff.

We argue that tool-making — creating reusable solver components, formulation templates, or heuristic algorithms — is the most promising direction. If the upfront cost of discovering a reusable tool is less than the cumulative cost of per-instance reasoning, tool creation is the better investment. This threshold is easily met for large problem families, as demonstrated by the LATM framework \cite{cai2024latm}, which enables unbounded subsequent lightweight-model calls with a single powerful-model creation pass. As models grow stronger in capability, the per-instance opportunity cost of direct solving increases, making the case for tool-creating approaches stronger over time.

\bibliographystyle{IEEEtran}
\bibliography{references}

\end{document}